\documentclass[a4paper,fleqn]{cas-dc}

\usepackage[authoryear,longnamesfirst]{natbib}
\usepackage{graphicx}
\usepackage{algorithm}
\usepackage{algorithmic}
\usepackage{booktabs}
\usepackage{multirow}
\usepackage{graphicx}
\usepackage{subfigure}
\def\tsc#1{\csdef{#1}{\textsc{\lowercase{#1}}\xspace}}
\tsc{WGM}
\tsc{QE}

\begin{document}
\let\WriteBookmarks\relax
\def\floatpagepagefraction{1}
\def\textpagefraction{.001}
\let\printorcid\relax 
\shorttitle{Progressive Fine-to-Coarse Reconstruction}

\shortauthors{Rui Ding et al}

\title [mode = title]{Progressive Fine-to-Coarse Reconstruction for Accurate Low-Bit Post-Training Quantization in Vision Transformers}

\author[label1]{Rui Ding}
\credit{Writing the original draft, Review \& editing, Visualization, Validation, Software, Methodology, Conceptualization}
\ead{dingrui961210@126.com}
\author[label1]{Liang Yong}
\credit{Validation, Software}
\ead{202212131152@stu.cqu.edu.cn}

\author[label1]{Sihuan Zhao}
\credit{Validation, Software}
\ead{17762330892@163.com}

\author[label1]{Jing Nie}
\credit{Validation, Software}
\ead{jingnie@cqu.edu.cn}

\author[label1]{Lihui Chen}
\credit{Review \& editing}
\ead{lihuichen@126.com}

\author[label1]{Haijun Liu}
\credit{Supervision, Review \& editing}
\ead{haijun_liu@126.com}

\author[label1]{Xichuan Zhou\corref{cor1}}
\credit{Supervision, Resources, Funding acquisition}
\ead{zxc@cqu.edu.cn}

\cortext[cor1]{Corresponding author at School of Microelectronics and Communication Engineering, Chongqing University, Chongqing, 401331, China; Email: zxc@cqu.edu.cn}
\affiliation[label1]{School of Microelectronics and Communication Engineering, Chongqing University, Chongqing, 401331, China}

\tnotemark[1]

\tnotetext[1]{This work was supported by the National Natural Science Foundation of China (U2133211, 62301092, 62301093).}


















\begin{abstract}
Due to its efficiency, Post-Training Quantization (PTQ) has been widely adopted for compressing Vision Transformers (ViTs). However, when quantized into low-bit representations, there is often a significant performance drop compared to their full-precision counterparts. To address this issue, reconstruction methods have been incorporated into the PTQ framework to improve performance in low-bit quantization settings. Nevertheless, existing related methods predefine the reconstruction granularity and seldom explore the progressive relationships between different reconstruction granularities, which leads to sub-optimal quantization results in ViTs.
To this end, in this paper, we propose a Progressive Fine-to-Coarse Reconstruction (PFCR) method for accurate PTQ, which significantly improves the performance of low-bit quantized vision transformers. Specifically, we define multi-head self-attention and multi-layer perceptron modules along with their shortcuts as the finest reconstruction units. After reconstructing these two fine-grained units, we combine them to form coarser blocks and reconstruct them at a coarser granularity level. We iteratively perform this combination and reconstruction process, achieving progressive fine-to-coarse reconstruction. Additionally, we introduce a Progressive Optimization Strategy (POS) for PFCR to alleviate the difficulty of training, thereby further enhancing model performance.
Experimental results on the ImageNet dataset demonstrate that our proposed method achieves the best Top-1 accuracy among state-of-the-art methods, particularly attaining 75.61\% for 3-bit quantized ViT-B in PTQ. Besides, quantization results on the COCO dataset reveal the effectiveness and generalization of our proposed method on other computer vision tasks like object detection and instance segmentation.
\end{abstract}





\maketitle

\section{Introduction}
In recent years, Vision Transformers (ViTs) \cite{ViToridosovitskiy2021an} have attracted increasing attention due to their impressive performance and have achieved great success in various categories of vision tasks, such as image classification \cite{NNClass1,NNClass2, CF-ViTchen2023cf}, semantic segmentation \cite{NNSeg1,NNSeg2, ViTseg1dong2023head}, object detection \cite{NNDetect1,NNDetect2, ViTobject1cao2022cf, ViTobject2zhu2021deformable}, and so on. The basic block within ViTs consists of the Multi-Head Self-Attention (MHSA) and the Multi-Layer Perceptron (MLP), where the former is used to model long-distance dependencies, and the latter is used to enhance feature representation capabilities \cite{ViToridosovitskiy2021an}. Built on multiple basic blocks with large feature dimensions, ViTs typically suffer from huge memory and computational requirements, posing deployment difficulties on resource-limited edge devices.

To address this problem, model quantization has been proposed and is considered a promising compression solution, which quantizes the full-precision weights and activations within ViTs into low-bit representations, significantly reducing memory and computational costs. This technique is commonly divided into two categories: Quantization-Aware Training (QAT) \cite{LSQEsser2020LEARNED, Quantizabletransformersbondarenko2024quantizable,QAT2BinaryViTle2023binaryvit} and Post-Training Quantization (PTQ) \cite{RepQli2023repq, FQ-ViTlin2021fq,PTQ1TCSVT10239177}. Unlike the costly QAT, which requires retraining on the entire dataset from scratch, PTQ only needs to calibrate the quantization parameters on a small-scale unlabeled dataset, making it suitable for efficiently quantizing ViTs. However, when quantized into low-bit (under 6-bit), there exists an unacceptable performance drop compared with the full-precision counterparts, such as 8.51\% and 36.23\% accuracy loss in 4-bit and 3-bit quantized ViT-S, respectively \cite{Outlier-awarema2024outlieraware, IS-ViTzhong2023s}.
\begin{figure}[t]
\centering
\includegraphics[width=1\columnwidth]{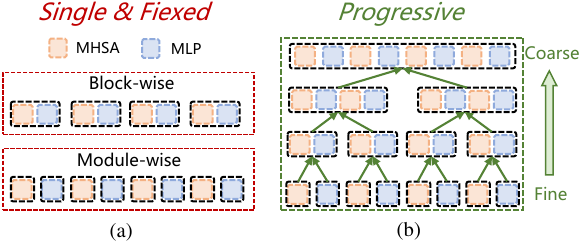} 
\caption{Comparisons between the conventional reconstruction methods with the single and fixed granularity and our proposed progressive fine-to-coarse reconstruction method.}
\label{fig1}
\end{figure}

To bridge the performance gap, reconstruction methods have been incorporated into PTQ for performance improvement with limited resource requirements, aiming to reduce the output difference between the quantized model and the full-precision model. As shown in Figure \ref{fig1}(a), conventional reconstruction granularity is single and fixed during optimization, such as module-wise and block-wise. For block-wise reconstruction, BRECQ \cite{BRECQli2021brecq} first attempts to reconstruct the quantized model in a block-wise manner, reducing generalization errors. PD-Quant \cite{PD-Quantliu2023pd} then improves the performance of block-wise reconstruction by introducing the prediction difference loss. A number of related works \cite{IS-ViTzhong2023s, ReconstructCLshang2024enhancing} follow this setting. In contrast, for module-wise reconstruction, Outlier-Aware \cite{Outlier-awarema2024outlieraware} proposes more reconstruction granularity using a fixed slicing strategy. ADFQ-ViT \cite{ADFQ-ViTjiang2024adfq} proposes a module-wise strategy to better minimize the quantization errors. However, these studies commonly apply the \textbf{single} reconstruction granularity and conduct the reconstruction process at this \textbf{fixed} granularity level. They ignore the \textbf{progressive} relationships between different reconstruction granularities where the finer reconstruction units are the building components for the coarser reconstruction units, which leads to inferior performance in low-bit quantized ViTs.

To this end, in this paper, we propose a Progressive Fine-to-Coarse Reconstruction (PFCR) for low-bit post-training quantization in vision transformers, aiming to bridge the performance gap between low-bit quantized ViTs and their full-precision counterparts. As illustrated in Figure \ref{fig1}(b), compared with conventional methods, the proposed PFCR reduces the output difference between the quantized models and full-precision models from fine granularity to coarse granularity progressively, optimizing different granularity levels during reconstruction and alleviating the performance degradation of low-bit quantized models significantly. Specifically, we first define the MHSA and MLP modules along with their shortcuts as the finest reconstruction units and granularity. Secondly, once their optimization is completed, we combine them to build the basic block and conduct the coarser block-wise reconstruction. Then, after block-wise optimization is finished, we conduct the same combination and reconstruction with two blocks at a coarser level. By progressively combining finer reconstructed units and reconstructing coarser units, the reconstruction is conducted iteratively and progressively from fine granularity to coarse granularity. Notably, the proposed PFCR has the following two advantages to benefit the reconstruction process: (1) Finer reconstruction granularity provides a better initialization for coarser reconstruction. (2) Coarser reconstruction in turn refines the parameters of finer granularity. Therefore, by applying this progressive optimization, the reconstruction errors could be significantly reduced, leading to the improved quantization performance.
Furthermore, to push the limits of low-bit quantized ViTs, we propose a Progressive Optimization Strategy (POS) designed for PFCR to achieve further accuracy improvement, where the reconstruction process is decomposed into two stages with different PFCR settings for better convergence and results.

In conclusion, our contributions in this paper can be summarized as follows:
\begin{itemize}
\item We propose PFCR, a novel progressive post-training quantization method for ViTs using fine-to-coarse reconstruction, which reduces the performance drop under low-bit quantization settings.
\item To further improve the performance of PFCR, we propose an effective reconstruction strategy, POS, which alleviates the optimization difficulty and pushes the limits of low-bit quantized ViTs.
\item Experimental results on the ImageNet dataset demonstrate that the proposed method achieves state-of-the-art performance under different quantization bit-width. Specifically, the 3-bit ViT-B achieves the Top-1 accuracy of 75.61\%, reducing the performance drop significantly.
\end{itemize}

\section{Related Works}
\subsection{Post-Training Quantization for ViTs}
Compared with the time-consuming Quantization-Aware Training (QAT) methods \cite{QAT1QuantFormerwang2022quantformer,QAT2BinaryViTle2023binaryvit,QAT3BiViThe2023bivit,NNQAT4CNN1}, which require retraining the whole model on the entire training dataset from scratch, Post-Training Quantization (PTQ) methods demonstrate superior efficiency by conducting calibration on small-scale unlabeled datasets. However, PTQ for vision transformers often suffers from an unacceptable accuracy drop compared to full-precision counterparts. Directly applying quantization methods designed for Convolutional Neural Networks (CNNs) \cite{LSQEsser2020LEARNED,PTQ4CNN1} in ViTs usually does not work well due to the unique structures like Multi-Head Self-Attention (MHSA), Multi-Layer Perceptron (MLP), and LayerNorm.

To alleviate this problem, numerous PTQ works dedicated to vision transformers have been proposed. Early works, such as FQ-ViT \cite{FQ-ViTlin2021fq}, propose power-of-two factor and log-int-softmax methods to systematically reduce the performance drop and sustain the non-uniform distribution, building a fully quantized ViT. Additionally, VT-PTQ \cite{VT-PTQliu2021post} incorporates a ranking loss into the objective function, aiming to retain the order of results from the self-attention module after quantization. In contrast, PTQ4ViT \cite{PTQ4ViTyuan2022ptq4vit} proposes a twin uniform quantization strategy to mitigate the errors induced by the abnormal activation distributions in ViTs. APQ-ViT \cite{APQ-ViTding2022towards} proposes a novel approach to optimize the calibration metric block-wise and introduces the Matthew-effect preserving quantization to retain the function of the attention mechanism. NoisyQuant \cite{NoisyQuantliu2023noisyquant} finds that for a given quantizer, adding a fixed uniform noisy bias to the values being quantized can significantly reduce the quantization error. Based on this observation, NoisyQuant first achieves the active modification of the heavy-tailed activation distribution for better quantization performance. Furthermore, RepQ-ViT \cite{RepQli2023repq} decouples the quantization and inference processes, where the former enjoys the high performance of complex quantizers and the latter utilizes the scale reparameterization for hardware-friendly acceleration.

\subsection{Reconstruction in Post-Training Quantization}
Although specially designed PTQ methods for ViTs improve the model performance to some extent, there still exists a notable accuracy drop compared to full-precision models in low-bit quantization scenarios. For example, the 4-bit quantized ViT-S using PTQ method RepQ-ViT \cite{RepQli2023repq} suffers a 16.34\% Top-1 accuracy drop on the ImageNet dataset.

To address this issue, reconstruction methods have been incorporated into the PTQ framework, aiming to further boost the capacity of quantized models by minimizing the output difference between the quantized and full-precision models. BRECQ \cite{BRECQli2021brecq} first proposes reconstructing the quantized model at a block-wise granularity level and identifies the overfitting problem of network-wise reconstruction, which shows superior effectiveness on CNNs. However, integrating BRECQ into PTQ for ViTs does not bring significant performance improvements in low-bit quantization. Therefore, PD-Quant \cite{PD-Quantliu2023pd} introduces the prediction difference loss from the network-wise level to the block-wise reconstruction which is aligned with the objective function, demonstrating better generalization performance for image classification. Besides, ADFQ-ViT \cite{ADFQ-ViTjiang2024adfq} presents attention-score-enhanced module-wise optimization for finer reconstruction, which is proven to reduce quantization errors. Furthermore, Outlier-Aware \cite{Outlier-awarema2024outlieraware} first introduces the concept of reconstruction granularity and demonstrates the influence of different granularities from the perspective of reducing outlier problems. It is worth noting that the difference from our work is that Outlier-Aware still applies the single reconstruction granularity by slicing the basic block and does not consider the progressive relationships among them.
\begin{figure*}[t]
\centering
\includegraphics[width=0.9\textwidth]{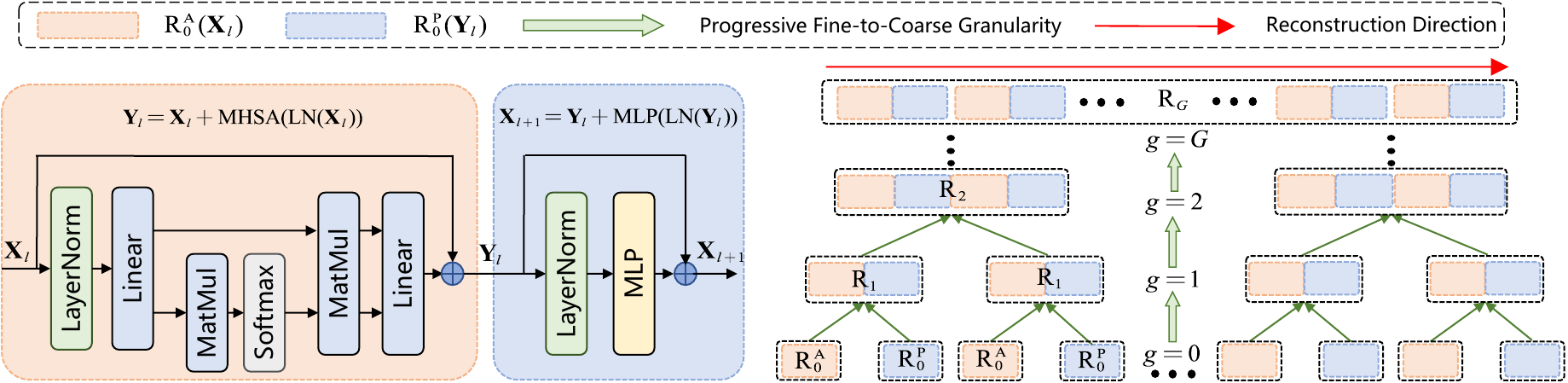} 
\caption{Illustration of the proposed PFCR. The left sub-figure shows the two finest reconstruction units with granularity level $g=0$. The right sub-figure demonstrates the iterative process of reconstruction and combination from fine granularity to coarse granularity progressively.}
\label{fig2}
\end{figure*}

\section{Method}
\subsection{Preliminaries}
\subsubsection{Modules in ViTs.} For a single vision transformer, the input image is first decomposed into $N$ flattened image patches and then projected to feature matrix represented by $\mathbf{X}_0 \in \mathbb{R}^{N \times D}$. The processed features are then fed into $L$ transformer basic blocks, which contain the Multi-Head Self-Attention (MHSA), Multi-Layer Perceptron (MLP) and LayerNorm (LN). The forward computation process of the $l$-th basic block can be formulated as follows:
\begin{align}
&\mathbf{Y}_{l}=\mathbf{X}_{l} + {\rm MHSA}({\rm LN}(\mathbf{X}_{l})), \label{mhsa} \\
&\mathbf{X}_{l+1}=\mathbf{Y}_{l} + {\rm MLP}({\rm LN}(\mathbf{Y}_{l})), \label{mlp}
\end{align}
where both the MHSA and MLP utilize the shortcuts to enhance the representational capability and improve the training performance.

\subsubsection{Quantizers in ViTs.}
Following the common strategies in recent works \cite{RepQli2023repq, IGQ-ViTmoon2024instance, IS-ViTzhong2023s}, we adopt the uniform quantization for quantizing all the weights and most of the activations within ViTs. In detail, considering $b$-bit quantization, a full-precision $x$ can be uniformly mapped to a integer $x_{\rm q}$ as follows:
\begin{equation}
    x_{\rm q} = {\rm clamp}(\lfloor \frac{x}{s} \rceil + z, 0, 2^b -1),
    \label{uniform}
\end{equation}
where $\lfloor \cdot \rceil$ is the rounding operation, and the clamp function limits the quantized value within the representation range from 0 to $2^b -1$. Besides, the $s$ and $z$ denote the quantization scale and zero-point which can be calibrated through the PTQ process on the small-scale calibration dataset as follows:
\begin{equation}
    s=\frac{{\rm max}(x)-{\rm min}(x)}{2^b-1},z=\lfloor -\frac{{\rm min}(x)}{s} \rceil,
    \label{scale_zeropoint}
\end{equation}
where the dequantization process can be formulated as:
\begin{equation}
    x_{\rm deq} = s \cdot (x_{\rm q}-z) \approx x.
\end{equation}
In particular, the softmax operation converts the attention scores within MHSA into probabilities named post-softmax activations, which follow the power-law distribution. Therefore, we apply the commonly used $\rm log_2$ quantizer for them as follows:
\begin{equation}
    x_{\rm q} = {\rm clamp}(\lfloor -{\rm log}(\frac{x}{s}) \rceil, 0, 2^b -1).
    \label{pot}
\end{equation}
The de-quantization value is computed by:
\begin{equation}
    x_{\rm deq} = s \cdot 2^{x_{\rm q}},
\end{equation}
which can be implemented with the efficient bit-shifting operations \cite{RepQli2023repq} for supporting the fast model inference.

During the optimization process, by utilizing the Straight Through Estimator (STE) \cite{STEbengio2013estimating} to propagate the gradients through the rounding function $\lfloor \cdot \rceil$, weights and quantization parameters in quantized ViTs can be updated through back-propagation algorithms.

\subsection{Progessive Fine-to-Course Reconstruction}
Reconstruction methods aim to restore performance by decreasing the output difference between quantized models and full-precision counterparts. Conventional works usually predefine the reconstruction granularity before the optimization process, such as module-wise and block-wise. However, few studies explore the progressive relationship between these granularity levels, where the finer reconstruction units are the building components for the coarser ones.

Based on the above observations, we propose a novel reconstruction method for ViTs by utilizing the internal connections between different granularity levels, named Progressive Fine-to-Coarse Reconstruction (PFCR). As shown in Figure \ref{fig2} and defined in Eq. (\ref{mhsa}) along with Eq. (\ref{mlp}), we choose the MHSA and MLP with their shortcuts as our finest reconstruction units, which can be represented with:
\begin{align}
{\rm R_{0}^{\rm A}}(\mathbf{X}_{l}) &=\mathbf{X}_{l} + {\rm MHSA}({\rm LN}(\mathbf{X}_{l})), \label{mhsa} \\
{\rm R_{0} ^{\rm P}}(\mathbf{Y}_{l}) &= \mathbf{Y}_{l} + {\rm MLP}({\rm LN}(\mathbf{Y}_{l})), \label{mlp}
\end{align}
where we consider ${{\rm R}_{g}(\cdot)}$ function as the reconstruction units, the superscript $\rm A$ and $\rm P$ denote the two finest reconstruction units, and the subscript $g=0$ is the finest reconstruction granularity level. Then we define the value range of $g$ as:
\begin{equation}
    g = \{0, 1, 2, ..., G\}, \label{GL}
\end{equation}
where there are total $G+1$ granularity levels and start from index 0. The bigger number of $g$ means the coarser reconstruction granularity and vice versa, therefore $G$ represents the coarsest reconstruction granularity.

The reason why we define ${\rm R_{0}^{\rm A}}(\mathbf{X}_{l})$ and ${\rm R_{0} ^{\rm P}}(\mathbf{Y}_{l})$ as the finest reconstruction granularities can be summarized into three folds: (1) the finer granularity reconstruction will consume more training time, damaging the efficiency of PTQ, (2) these two units retain structural and functional integrity, (3) the shortcuts in these modules provide effective gradient flows for better optimization results.

Therefore, as illustrated in the right sub-figure of Figure \ref{fig2}, the PFCR begins from these two finest units at the granularity level $g=0$, and the minimizing objective function is formulated as:
\begin{align}
       &{\rm min} \| {\rm R}_{0}^{\rm A}(\mathbf{X}_l)-\hat{\rm R}_{0}^{\rm A}(\mathbf{X}_l) \|_{2}^2, \label{loss-g0-1} \\
      &{\rm min}  \| {\rm R}_{0}^{\rm P}(\mathbf{Y}_l)-\hat{\rm R}_{0}^{\rm P}(\mathbf{Y}_l) \|_{2}^2, \label{loss-g0-2}
\end{align}
where we use Mean Square Error (MSE) denoted by $\| \cdot\|_{2}^2$ to measure the output errors between the quantized units $\hat{\rm R}$ and the full-precision counterparts $\rm R$. We adopt the prevalent optimization algorithms to update the weights of quantizers in ViTs for minimizing the object function. Due to the simplified module structure, the over-fitting problem will be mitigated during the reconstruction process \cite{BRECQli2021brecq}.

After they are reconstructed, we combine these two fine units to build the reconstruction targets ${\rm R}_{1}$ with a coarser granularity level $g = 1$. At this level, we conduct the block-wise reconstruction similar to other works \cite{IS-ViTzhong2023s, BRECQli2021brecq, Outlier-awarema2024outlieraware}, as follows:
\begin{equation}
   {\rm min} \| {\rm R}_{1}(\mathbf{X}_l)-\hat{\rm R}_{1}(\mathbf{X}_l) \|_{2}^2,
    \label{g1}
\end{equation}
where ${\rm R}_{1}$ and $\hat{\rm R}_{1}$ denote the full-precision and quantized blocks respectively, which are built on the units from finer granularity level $g=0$. In this case, with Eq. (\ref{mhsa}) and Eq. (\ref{mlp}), we could rewrite the Eq. (\ref{g1}) as follows:
\begin{equation}
 {\rm min}    \| {\rm R}_{0}^{\rm P}({\rm R}_{0}^{\rm A}(\mathbf{X}_l))-\hat{\rm R}_{0}^{\rm P}(\hat{\rm R}_{0}^{\rm A}(\mathbf{X}_l)) \|_{2}^2.
    \label{loss-g1}
\end{equation}
where the coarse block-wise reconstruction is decomposed into two finer module-wise reconstructions that have been optimized in the last granularity level ($g=0$) with Eq. (\ref{loss-g0-1}) and Eq. (\ref{loss-g0-2}). By inheriting the optimized weights from the finer level, this strategy provides a better \textbf{initialization} for the current reconstruction granularity and the coarser reconstruction \textbf{finetunes} the parameters further, boosting the converge speed and reducing the reconstruction loss.

As shown in Figure \ref{fig2}, the coarser reconstruction for $ {\rm R}_{g}$ will be conducted when two finer units $ {\rm R}_{g-1}$ reconstructed, therefore the coaserest granularity level $G$ is computed by:
\begin{equation}
G=\begin{cases}
 {\rm log}_{2}(2L), &{\rm if} \; (2L\,\& \,(2L-1))=0,\\
 \lfloor {\rm log}_{2}(2L) \rceil-1, &{\rm else},
\end{cases}
    \label{G}
\end{equation}
where $\&$ denotes the bit-wise AND operation, the $2L$ indicates the total number of finest reconstructions units in $L$ blocks for ViTs. For some cases where the number is not a power of 2, we apply the rounding operation and ignore the last granularity level. Taking the 12-block ViT \cite{ViToridosovitskiy2021an} as an example, the coarsest granularity level $G=3$  is computed by Eq. (\ref{G}), where the $2^0$, $2^1$, $2^2$ and $2^3$ number of finest units are assigned for building the different reconstruction granularity from $g=0$ to $g=3$.

Then the general object function of PFCR for the $l$-th block can be formulated as:
\begin{equation}
    {\rm min}\| {\rm R}_{g}(\mathbf{X}_g)-\hat{\rm R}_{g}(\mathbf{X}_g) \|_{2}^2, \label{object-function}
\end{equation}
where $g = \{0, 1, 2, ..., G\}$ and ${\rm R}_{g}(\mathbf{X}_g)$ is derived by:
\begin{equation}
\begin{cases}
 {\rm R}_{g-1}({\rm R}_{g-1}(\mathbf{X}_{l-2^{g-1}+1})), &(2l \, \% \,2^{g}) = 0 \land 1<g\leq G, \\
 {\rm R}_{0}^{\rm P}({\rm R}_{0}^{\rm A}(\mathbf{X}_l)), & g = 1, \\
{\rm R}_{0}^{\rm P}(\mathbf{Y}_l) \;{\rm or} \;{\rm R}_{0}^{\rm A}(\mathbf{X}_l),  &g=0.
 \end{cases}
 \label{Rg}
\end{equation}
Conducting the process with Eq. (\ref{object-function}) iteratively, we reconstruct the whole model from fine to coarse progressively, which reduces the reconstruction loss significantly and contributes to improved quantization performance.


\subsection{Progressive Optimization Strategy}
Based on the observations from I\&S-ViT \cite{IS-ViTzhong2023s}, the loss landscape of ViTs with quantized weights and activations is rugged and drastically changing, resulting in poor training performance. Previous works like ReAct \cite{ReActNetliu2020reactnet} propose a two-stage framework equipped with distributional loss to build high-performance binary neural networks, and I\&S-ViT also introduces the three-stage method for better results.

\begin{algorithm}[t]
\caption{Progressive Optimization Strategy for PFCR}
\label{POS_algo}
\textbf{Input}: Pretrained full-precision model $\rm F$, calibration data $\mathcal{C}$\\
\textbf{Parameter}: Base iteration $iter_0$, base learning rate $lr_0$, block number $L$, coarsest granularity level $G$, quantization bit-width $b$
\begin{algorithmic}[1]
\STATE \textbf{Stage1}: Quantize the activations of $\rm F$ to $b$-bit by Eq. (\ref{uniform}) and Eq. (\ref{pot}) using $\mathcal{C}$. Set $G=1$, $iter_g$ and $lr_g$ with Eq. (\ref{lr}) and Eq. (\ref{iter}).
\FOR{$g=0,1$}
\FOR{$l=1,2,...,L$}
\STATE Reconstruct ${\rm R}_{g}$ with Eq. (\ref{object-function}-\ref{Rg}) for $iter_g$.
\ENDFOR
\ENDFOR
\STATE \textbf{Stage2}: Quantize the activations and weights of $\rm F$ to $b$-bit by Eq. (\ref{uniform}) and Eq. (\ref{pot}) with $\mathcal{C}$. Set $G$ with Eq. (\ref{G}), $iter_g$ and $lr_g$ with Eq. (\ref{lr}) and Eq. (\ref{iter}).
\FOR{$g=0,1,...,G$}
\FOR{$l=1,2,...,L$}
\STATE Reconstruct ${\rm R}_{g}$ with Eq. (\ref{object-function}-\ref{Rg}) for $iter_g$.
\ENDFOR
\ENDFOR
\end{algorithmic}
\textbf{Output}: Quantized model $\rm Q$ with $b$-bit representation
\end{algorithm}
In this paper, to push the limit of the low-bit quantized ViTs, we propose a Progressive Optimization Strategy (POS), combining the two-stage progressive optimization with the progressive fine-to-coarse reconstruction. Specifically, the process can be summarized as follows:

\textbf{(1) Smooth Training with Less Granularity.} In the first stage, we retain the weights of ViTs in full-precision representation while quantizing the activations into the target low-bit bit-width. This method could provide a smoother and less variation loss landscape for the optimization process, which alleviates the training challenge for reconstruction. Due to this great property, we set $G=1$ in this stage to balance the trade-off between performance and efficiency.

\textbf{(2) Rugged Training with More Granularity.} In the second stage, we quantize both the weights and activations in ViTs into low-bit representations, where the loss landscape of reconstruction becomes rugged and varied. To alleviate the training difficulties, we compute the coarsest granularity by Eq. (\ref{G}). Combining the fine-to-coarse method and inheriting the weights from the first stage, the POS converges to better optimization results. The simplified structure of finer reconstruction units acts like a regularization for coarser-grained reconstruction, reducing the generalization error compared with the previous works with fixed granularity \cite{BRECQli2021brecq, IS-ViTzhong2023s, Outlier-awarema2024outlieraware}.

Given the unequal network capacity in various reconstruction granularity, we adopt the diminishing learning rates and incremental training iterations as follows:
\begin{align}
    lr_{g} &= lr_0 * (1-0.2*g), \label{lr}\\
    iter_{g} &= iter_0 * (1+0.2*g), \label{iter}
\end{align}
where $lr_0$ and $iter_0$ denote the base learning rate and iteration in $g=0$, respectively. The base iteration number impacts the reconstruction time and the quantization performance in low-bit quantization, and we will investigate the effect in detail within the experimental section.

With PFCR and POS, we provide the optimization pipeline of the reconstruction method in Algorithm \ref{POS_algo}.
\begin{table*}[t]
\setlength{\tabcolsep}{2.7mm}
\centering
\caption{Comparisons of quantization results with other State-Of-The-Art (SOTA) post-quantization methods on the ImageNet dataset using different vision transformer structures and various low-bit quantization settings. The Top-1 accuracy (\%) is reported as the performance metric. ``W/A'' denotes the quantization bit-width for weights and activations respectively. }
\begin{tabular}{cccccccc}
\toprule
\textbf{Method}  & \textbf{Bit-width (W/A)} & \textbf{ViT-S} & \textbf{ViT-B} & \textbf{DeiT-S} & \textbf{DeiT-B} & \textbf{Swin-S} & \textbf{Swin-B} \\
\midrule
Full-Precision   & 32/32                    & 81.39          & 84.54                & 79.85           & 81.80           & 83.23           & 85.27           \\ \midrule
BRECQ  \cite{BRECQli2021brecq}           & 3/3                      & 0.42           & 0.59             & 14.63           & 46.29           & 11.67           & 1.70            \\
PD-Quant \cite{PD-Quantliu2023pd}           & 3/3                      & 1.77           & 13.09          &29.33           & 0.94            & 69.67           & 64.32           \\
I\&S-ViT  \cite{IS-ViTzhong2023s}              & 3/3                      & 45.16          & 63.77      & 55.78           & 73.30          & 74.20     & 69.30    \\
Ours                   & 3/3                      & \textbf{58.94}             & \textbf{75.61}                & \textbf{65.70}              & \textbf{75.25}             & \textbf{75.78}           & \textbf{70.87}              \\
\midrule
PTQ4ViT  \cite{PTQ4ViTyuan2022ptq4vit}                & 4/4                      & 42.57          & 30.69            & 34.08           & 64.39           & 76.09           & 74.02           \\
APQ-ViT \cite{APQ-ViTding2022towards} & 4/4 & 47.95 & 41.41 & 43.55& 67.48 & 77.15 & 76.48 \\
BRECQ   \cite{BRECQli2021brecq}                  & 4/4                      & 12.36          & 9.68              & 63.73           & 72.31           & 72.74           & 58.24           \\
PD-Quant \cite{PD-Quantliu2023pd}                 & 4/4                      & 1.51           & 32.45            & 71.21           & 73.76           & 79.87           & 81.12           \\
Repq-ViT \cite{RepQli2023repq}                  & 4/4                      & 65.05          & 68.48           & 69.03           & 75.61           & 79.45           & 78.32           \\
I\&S-ViT   \cite{IS-ViTzhong2023s}                & 4/4                      & 74.87          & 80.07           & 75.81           & 79.97           & 81.17           & 82.60           \\
ADFQ-ViT \cite{ADFQ-ViTjiang2024adfq} & 4/4 &72.14 & 78.71 & 75.06& 78.75 & 80.63& 82.33\\
Outlier-Aware \cite{Outlier-awarema2024outlieraware}             & 4/4                      & 72.88          & 76.59             & 76.00           & 78.83           & 81.02           & 82.46           \\
Ours              & 4/4                      & \textbf{76.31}            & \textbf{83.19}            & \textbf{76.24}              & \textbf{80.47}           & \textbf{81.54}              & \textbf{82.85}             \\
\midrule
PTQ4ViT    \cite{PTQ4ViTyuan2022ptq4vit}               & 6/6                      & 78.63          & 81.65            & 76.28           & 80.25           & 82.38           & 84.01           \\
APQ-ViT \cite{APQ-ViTding2022towards} & 6/6 & 79.10 & 82.21 & 77.76& 80.42 & 82.67 & 84.18 \\
BRECQ  \cite{BRECQli2021brecq}                    & 6/6                      & 54.51          & 68.33                 & 78.46           & 80.85           & 82.02           & 83.94           \\
PD-Quant  \cite{PD-Quantliu2023pd}                  & 6/6                      & 70.84          & 75.82            & 78.40           & 80.52           & 82.51           & 84.32           \\
I\&S-ViT  \cite{IS-ViTzhong2023s}                  & 6/6                      & 80.43          & 83.82              & 79.15           & 81.68       & 82.89           & 84.94           \\
ADFQ-ViT \cite{ADFQ-ViTjiang2024adfq} & 6/6 &80.54 & 83.92 & 79.04& 81.53 & 82.81& 84.82\\
Outlier-Aware \cite{Outlier-awarema2024outlieraware}              & 6/6                      & 80.60          & 83.81               & \textbf{79.50}           & \textbf{81.72}           & 82.76           & 84.91           \\
Ours                      & 6/6                      & \textbf{80.64}            & \textbf{84.85}                   & 79.34             & 81.65              & \textbf{82.90}              & \textbf{84.98}              \\
\bottomrule
\end{tabular}
\label{sota}
\end{table*}

\section{Experiments}
\subsection{Experimental Settings}
\subsubsection{Models and Datasets} We apply the proposed PFCR along with POS to quantize different types of vision transformers, including ViT \cite{ViToridosovitskiy2021an}, DeiT \cite{DeiTtouvron2021training} and Swin-Transformer \cite{SwinTransliu2021swin} with small and base model scales. The quantization bit-width of these ViTs is set as 3-bit, 4-bit, and 6-bit for both weights and activations. The pre-trained models are adopted from Timm \cite{timmrw2019timm}.

After that, we evaluate the reconstruction performance on the validation set of ImageNet \cite{ImageNetdeng2009imagenet} which is a large-scale benchmark dataset for image classification with 1000 classes. The Top-1 accuracy is reported as the evaluation metric.

Besides, for validating the generalization of our proposed method, we evaluate the proposed PFCR on the COCO \cite{COCOlin2014microsoft} dataset and use the Mask R-CNN \cite{MaskRCNNhe2017mask} and Cascade Mask R-CNN \cite{CascadeRCNNcai2018cascade} as the frameworks for object detection and instance segmentation, respectively. We apply the Swin-Transformer \cite{SwinTransliu2021swin} as our backbone model and use the box and mask average precision as the evaluation metrics.
\subsubsection{Implementation Details} We implement our proposed PFCR and POS based on the Pytorch \cite{Pytorchpaszke2019pytorch} framework and conduct all the experiments on a single A6000 GPU. For calibration, we adopt the 64 randomly selected samples from the training set in ImageNet to initialize the quantization parameters in Eq. (\ref{uniform}) and Eq. (\ref{pot}). All the weights and activations are quantized into the target low-bit representations, including the output from Softmax attention and LayerNorm layer. Channel-wise uniform quantization is opted for weights, while layer-wise uniform quantization is used for activations except for the post-softmax activations that adopt the $\rm log_2$ quantizer.

For the ImageNet dataset, we randomly select 1024 samples for reconstruction. The default learning rate $lr_0$ is set as $4e-5$ and iteration number $iter_0$ is set as 800, 300, and 100 for 3-bit, 4-bit, and 6-bit quantization, respectively.

For the COCO dataset, we select only 1 sample for reconstruction, which is 1024$\times$ fewer than the previous method \cite{IS-ViTzhong2023s}. The default learning rate and iteration number are set as 6e-7 and 500 respectively, and the batch size is set as 1. Furthermore, the Adam optimizer \cite{Adamkingma2014adam} and cosine annealing scheduler are applied for updating parameters and adjusting the learning rate, respectively.

\begin{table*}[t]
\setlength{\tabcolsep}{1.5mm}
\centering
\caption{Comparisons of quantization results with other State-Of-The-Art (SOTA) post-quantization methods on the COCO dataset using different vision transformer structures such as Swin-T and Swin-S. ``W/A'' denotes the quantization bit-width for weights and activations respectively.``$\rm AP^{box}$'' denotes the box average precision (\%) for evaluating the performance of object detection while ``$\rm AP^{mask}$'' denotes the mask average precision (\%) for instance segmentation. $^*$ denotes that the results are reproduced by the official code.  }
\begin{tabular}{cccccccccc}
\toprule
\multicolumn{1}{c}{\multirow{3}{*}{\textbf{Method}}} & \multicolumn{1}{c}{\multirow{3}{*}{\textbf{Bit-width(W/A)}}} & \multicolumn{4}{c}{\textbf{Mask R-CNN}}                          & \multicolumn{4}{c}{\textbf{Cascade Mask R-CNN}}                  \\ \cline{3-10}
\multicolumn{1}{c}{}                        & \multicolumn{1}{c}{}                                & \multicolumn{2}{c}{\textbf{Swin-T}} & \multicolumn{2}{c}{\textbf{Swin-S}} & \multicolumn{2}{c}{\textbf{Swin-T}} & \multicolumn{2}{c}{\textbf{Swin-S}} \\
\multicolumn{1}{c}{}                        & \multicolumn{1}{c}{}                                & $\rm AP^{box}$       & $\rm AP^{mask}$      & $\rm AP^{box}$       & $\rm AP^{mask}$       & $\rm AP^{box}$       & $\rm AP^{mask}$       & $\rm AP^{box}$        & $\rm AP^{mask}$      \\ \hline
Full-Precision                              & 32/32                                               & 46.0        & 41.6         & 48.5        & 43.3         & 50.4        & 43.7         & 51.9         & 45.0        \\ \hline
PTQ4ViT  \cite{PTQ4ViTyuan2022ptq4vit}                                   & 4/4                                                 & 6.9         & 7.0          & 26.7        & 26.6         & 14.7        & 13.5         & 0.5          & 0.5         \\
APQ-ViT \cite{APQ-ViTding2022towards}                                     & 4/4                                                 & 23.7        & 22.6         & \textbf{44.7}        & 40.1         & 27.2        & 24.4         & 47.7         & 41.1        \\
BRECQ \cite{BRECQli2021brecq}& 4/4                                                 & 25.4        & 27.6         & 34.9        & 35.4         & 41.2        & 37.0         & 44.5         & 39.2        \\
QDrop \cite{Qdropwei2022qdrop}                                      & 4/4                                                 & 12.4        & 12.9         & 42.7        & 40.2         & 23.9        & 21.2         & 24.1         & 21.4        \\
PD-Quant \cite{PD-Quantliu2023pd}                                    & 4/4                                                 & 17.7        & 18.1         & 32.2        & 30.9         & 35.5        & 31.0         & 41.6         & 36.3        \\
RepQ-ViT \cite{RepQli2023repq}                                   & 4/4                                                 & 36.1        & 36.0         & 42.7$^*$        & 40.1$^*$         & 47.0        & 41.4         & 49.3         & 43.1        \\
I\&S-ViT  \cite{IS-ViTzhong2023s}                                    & 4/4                                                 & 37.5        & 36.6         & 43.4        & 40.3         & 48.2        & 42.0         & \textbf{50.3}         & 43.6        \\
Ours                                        & 4/4           & \textbf{39.3}           & \textbf{38.6}          & 43.2           & \textbf{41.2}          & \textbf{48.3}           & \textbf{42.5}           & 50.2 &\textbf{43.9}         \\ \bottomrule
\end{tabular}
\label{COCO_results}
\end{table*}
\subsection{Comparisons with SOTA on ImageNet dataset}
In Table \ref{sota}, we compare our method with other State-Of-The-Art (SOTA) works on ImageNet and report the Top-1 accuracy of various categories of quantized ViTs under different low-bit quantization settings, including 3-bit, 4-bit, 6-bit quantization for both weights and activations.

For 3-bit quantization, as shown in Table \ref{sota}, our proposed PFCR achieves superior PTQ performance based on different backbones, especially improving the Top-1 accuracy of 3-bit ViT-S and 3-bit ViT-B by 13.78\% and 11.84\% over the SOTA reconstrution method I\&S-ViT, respectively. In contrast, the block-wise methods BRECQ and PD-Quant with single and fixed reconstruction granularity suffer from a severe performance drop compared with the full-precision counterparts, from 13.56\% to 80.97\% in all the network structures. The significant performance improvement can be attributed to the more effective reconstruction process benefited from the progressive fine-to-coarse strategy combined with the POS framework.

For 4-bit quantization, our method still outperforms other SOTA post-training quantization methods based on all the vision transformer structures including ADFQ-ViT, I\&S-ViT and Outlier-Aware, improving the performance of 4-bit quantized ViTs by a considerable margin. To be specific, compared with the SOTA method Outlier-Aware which designs different reconstruction granularity for different models, our PFCR enjoys the benefits of progressive optimization from finer granularity to coarser granularity in a unified framework, obtaining significant performance boosting from 0.24\% to 6.60\% based on different ViTs. Furthermore, our proposed method significantly narrows the accuracy gap between the 4-bit quantized ViT-B and its full-precision counterpart to only 1.35\%. Similarly, the 4-bit quantized DeiT-B achieves 80.47\% Top-1 accuracy with merely a 1.33\% drop compared with its full-precision counterpart. These impressive results demonstrate the effectiveness of our proposed method.

For 6-bit quantization, the superiority of our proposed method becomes no longer such obvious and achieves the second-best results in some cases such as 81.65\% in 6-bit quantized DeiT-S, which is 0.07\% lower than the Outlier-Aware (81.72\%). This is because when the model capacity is large enough for effective reconstruction in relatively higher quantization bit-width, the proposed PFCR may cause an over-fitting problem on the small-scale calibration dataset by optimizing the object function, leading to extra generalization errors. In this case, we could consider a more elaborate hyper-parameter search or adopt the early-stopping technique. Despite this limit, our method achieves the SOTA results over most of the other PTQ methods. Notably, the 6-bit quantized ViT-B obtains the Top-1 accuracy of 84.85\%, which is even 0.31\% higher than the full-precision model.
\subsection{Comparisons with SOTA on COCO dataset}
To further explore the effectiveness of our proposed method on other high-level computer vision tasks, based on Mask-RCNN and Cascade Mask-RCNN frameworks, we conduct experiments on the COCO dataset and compare the 4-bit quantization results using Swin-T and Swin-S backbones in Table \ref{COCO_results}. The box average precision $\rm AP^{box}$ and mask average precision $\rm AP^{mask}$ are applied as the evaluation metrics for object detection and instance segmentation.

For the Mask R-CNN framework, as the result shows, the APQ-ViT achieves the best box average precision with the Swin-S backbone. Nevertheless, it suffers significant performance degeneration with the Swin-S backbone, which is shown that the box and mask average precision are 22.3\% and 19.0\% lower than the full-precision counterpart. In contrast, our proposed PFCR achieves consistent performance improvement in different backbone models (Swin-T and Swin-S) compared with the block-wise method such as BRECQ and PD-Quant, which demonstrates the effectiveness and robustness of the proposed progressive reconstruction technique. Particularly, when the Swin-T is used as the backbone, our proposed achieves 1.8 and 2.0 points higher average precision over the SOTA method I\&S-ViT for object detection and instance segmentation tasks respectively.

For the Cascade Mask R-CNN framework, our proposed PFCR obtains the best results on most evaluation metrics in various backbone models for both of the vision tasks. Specifically, When Cascade Mask R-CNN with Swin-T is used, our proposed PFCR improved the box AP and mask AP by 0.1 and 0.5 points over the previous SOTA method. When Swin-S serves as the backbone model, the segmentation performance of 4-bit quantized Cascade Mask R-CNN is further increased by 0.3 points in mask average precision. Meanwhile, the detection performance of our proposed method is comparable with the SOTA method. These results validate the effectiveness and generalization of our proposed reconstruction method on various high-level computer vision tasks.
\begin{table}[t]
\setlength{\tabcolsep}{2.2mm}
\centering
 \caption{Results of the ablation study on ImageNet in 4-bit quantized ViT-S. The Top-1 accuracy (\%) is reported. PFCR and POS refer to the proposed progressive fine-to-coarse reconstruction and progressive optimization strategy, respectively.}
\begin{tabular}{ccccc}
\toprule
Method  & W/A & PFCR & POS & Top-1 (\%) \\
\midrule
Full-Precision  & 32/32& $\times$  &  $\times$ &81.39\\
Block-wise  & 4/4 & $\times$  &  $\times$ &36.58 \\
\midrule
\multirow{3}{*}{\textbf{Ours}}  & 4/4 & \checkmark & $\times$ &74.03 \\
 & 4/4 &$\times$ & \checkmark & 75.25\\
  & 4/4 &\checkmark & \checkmark & \textbf{76.31}\\
\bottomrule
\end{tabular}
 \label{ablation_study_summary}
\end{table}

\begin{table}[t]
\setlength{\tabcolsep}{2.5mm}
\centering
\caption{Results of different reconstruction granularity $G$ in PFCR under 3-bit, 4-bit, and 6-bit quantization when trained for 250 iterations. The Top-1 accuracy (\%) is reported and ViT-S is applied as the quantized model. }
\begin{tabular}{c|ccccc}
\toprule
 Model & W/A & $G=0$ & $G=1$   & $G=2$   & $G=3$   \\ \midrule
 \multirow{3}{*}{\textbf{ViT-S}} &3/3     &  19.23   & 31.24   & 44.07 & 49.83 \\
  &4/4   &   15.96  & 26.25  & 42.44     &   51.53    \\
  &6/6    &   79.86  &    80.34 &    80.48   &    80.47   \\ \bottomrule
\end{tabular}
\label{PFCR_G}
\end{table}
\subsection{Ablation Study}
In this subsection, based on ViT-S, we provide a comprehensive study for analyzing the effect of the different components in our proposed method. We reproduce the block-wise reconstruction under the same setting and summarize the performance breakdown results in Table \ref{ablation_study_summary}, where the PFCR and POS respectively improve the quantization results by 37.45\% and 38.67\%, and equipped with both of them, the 4-bit quantized ViT-S achieves a preferable Top-1 accuracy of 76.31\%, which demonstrates the effectiveness and complementarity of our proposed methods for accurate low-bit quantization in ViTs.

\begin{figure}[t]
\centering
\includegraphics[width=0.75\columnwidth]{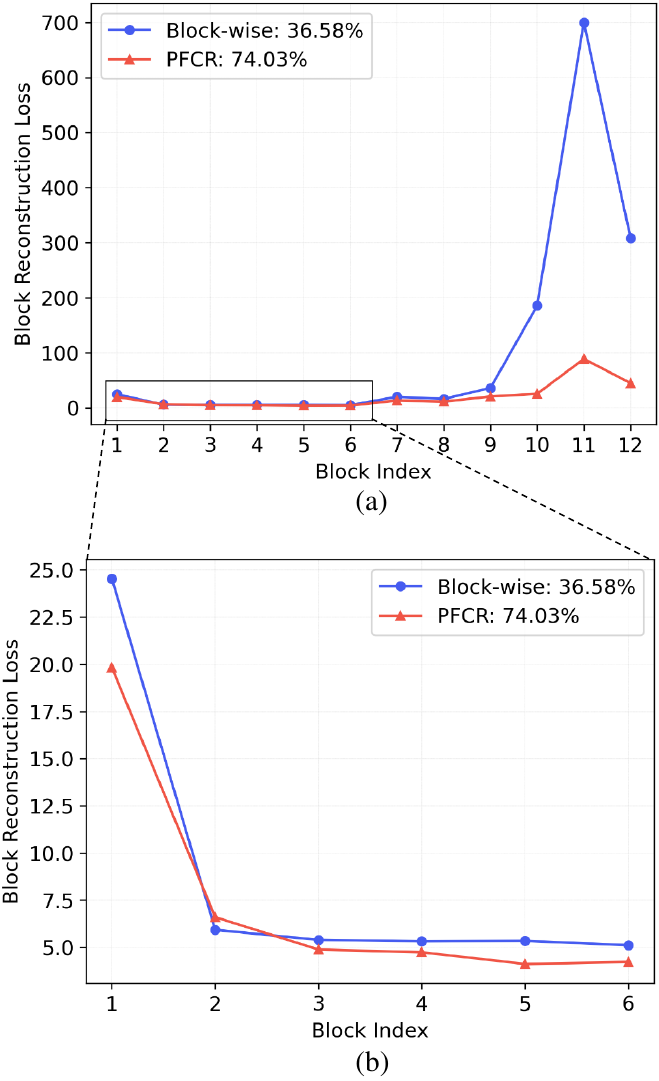}
\caption{Comparison of the reconstruction loss distribution in the 12-block 4-bit quantized ViT-S between conventional block-wise reconstruction and the proposed PFCR. (a) The overall loss distribution of 12 blocks in ViT-S. (b) The zoomed in loss distribution of the first 6 blocks in ViT-S.}
\label{pfcr_fig}
\end{figure}

\subsubsection{Effect of PFCR}
Firstly, we examine the effect of the proposed progressive fine-to-coarse reconstruction method.
Table \ref{PFCR_G} demonstrates the quantization results of ViT-S under different low-bit settings and reconstruction granularity levels defined by $G$. For this part in the Ablation Study, we only explore the effect of granularity $G$ on the quantization performance so the default iteration $iter_0=250$ is simply set as 250 iterations for experimental efficiency and eliminating the effect of longer training time. Besides, we use the one-stage reconstruction for PFCR to exclude the influence of POS.

As Table \ref{PFCR_G} shows, as $G$ grows larger, the performance of quantized ViT-S is improved significantly under 3-bit and 4-bit quantization settings, indicating the effectiveness of the proposed PFCR in low-bit quantization. For example, the 3-bit quantized ViT-S boosts its Top-1 accuracy from 19.23\% to 49.83\% by progressively reconstructing the model from fine-to-coarse granularity.  Nevertheless, the 6-bit quantized ViT-S suffers a 0.01\% accuracy drop when $G$ increases from 2 to 3, showcasing the over-fitting problem in higher-bit quantization which corresponds to the phenomenon in the previous section. Specifically, as revealed in Table \ref{ablation_study_summary}, compared with the normal block-wise reconstruction, our proposed PFCR achieves a significant 37.45\% accuracy improvement under the same experimental setting.

Furthermore, we explore the effect of PFCR in terms of the reconstruction loss which should be minimized during the optimization process. Figure \ref{pfcr_fig} provides the loss distribution of block-wise reconstruction and PFCR on 12-block ViT-S. It can be seen in Figure \ref{pfcr_fig}(a) that the loss of the block-wise method increases as the block index goes larger and explodes at the last three blocks, leading to severe performance degeneration. In contrast, the reconstruction loss of PFCR is relatively stable and flat across different block indexes, suppressing the accumulated quantization errors compared with the block-wise counterpart. Additionally, as illustrated in Figure \ref{pfcr_fig}(b), we zoom in on the loss distribution in the first several blocks and find that our proposed method achieves a consistently smaller reconstruction loss. These advantages are derived from the proposed PFCR, where the finer reconstruction provides a better initialization for coarser reconstruction and the coarser reconstruction in turn finetunes the parameters within the finer granularity. By conducting the reconstruction more stably and effectively, our proposed method boosts the performance of 4-bit quantized ViT-S from 36.58\% to 74.03\% as shown in Table \ref{ablation_study_summary}.

\begin{table}[t]
\setlength{\tabcolsep}{5mm}
\centering
\caption{Comparison of the quantization results between conventional one-shot reconstruction and the proposed two-stage POS framework. We report the Top-1 (\%) accuracy in three low-bit quantization bit-width and make the performance gain bold.}
\begin{tabular}{c|ccc}
\toprule
\textbf{ViT-S} & 3/3 & 4/4 & 6/6 \\ \midrule
One-stage     &  54.53   &74.03     &   80.46  \\
POS            &   58.94  &   76.31  &   80.64  \\ \midrule
Gain (\%)       &   \textbf{+4.41}  &  \textbf{+2.28}   &   \textbf{+0.18}  \\ \bottomrule
\end{tabular}
\label{POS}
\end{table}

\begin{table}[t]
\setlength{\tabcolsep}{0.3mm}
\centering
\caption{Detailed comparison of time costs and Memory requirements along with quantization performance between our method with the Outlier-Aware \cite{Outlier-awarema2024outlieraware} in 4-bit quantized ViT-S.}
\begin{tabular}{cccc}
\toprule
\textbf{Method}        & Top-1 (\%) & Time (Min) & Memory (GB)\\
\midrule
Outlier-Aware & 72.88      & 130.75  & 4.65        \\
PFCR w/o POS  & 74.03      & 14.13   & 11.02        \\
PFCR w/ POS   & 76.31      & 18.78  & 11.02  \\
\bottomrule
\end{tabular}
\label{timecost and top1}
\end{table}

\begin{figure}[ht]
\centering
\subfigure[Accuracy]{
\includegraphics[width=0.75\columnwidth]{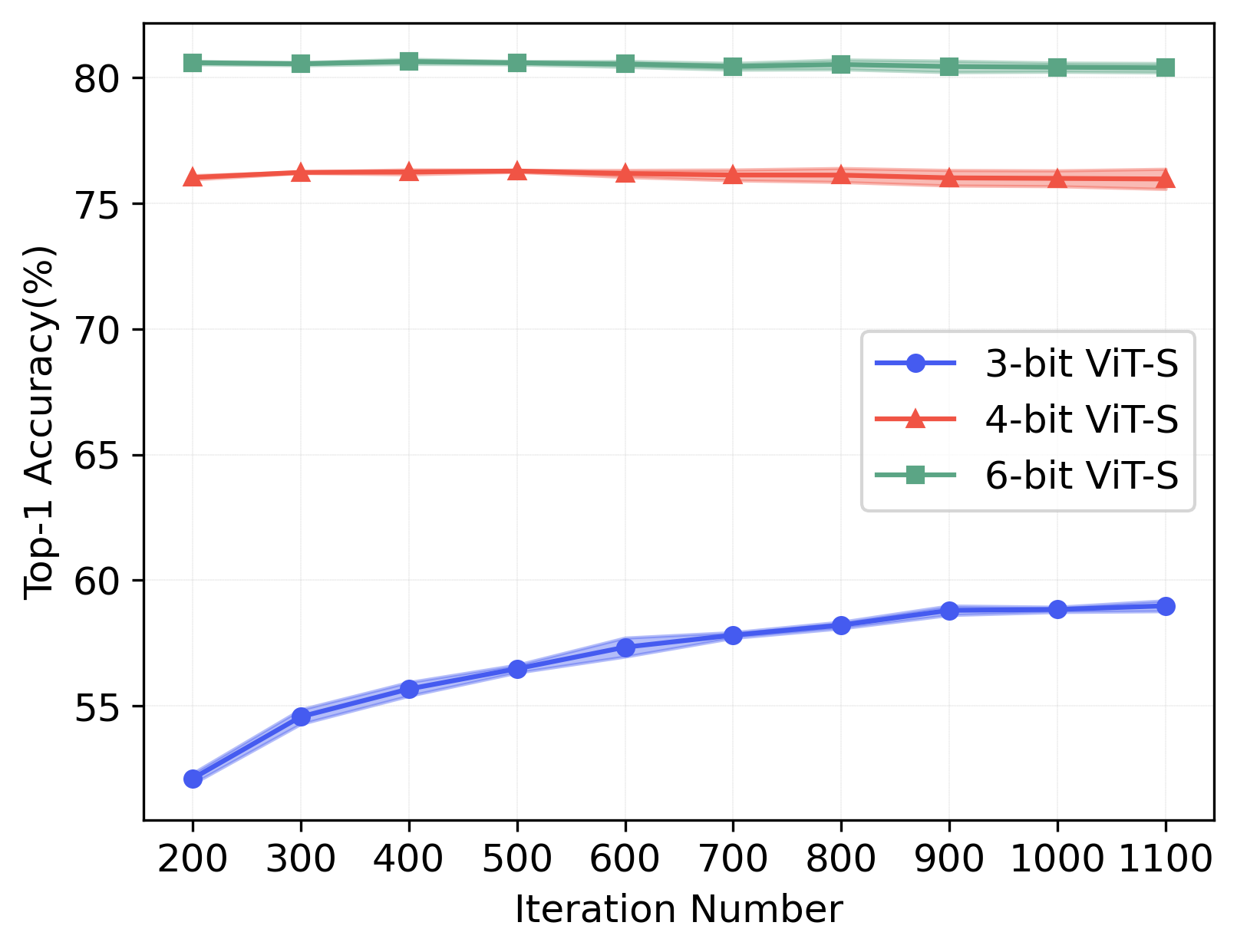}
\quad}
\\
\subfigure[Time cost]{
\includegraphics[width=0.75\columnwidth]{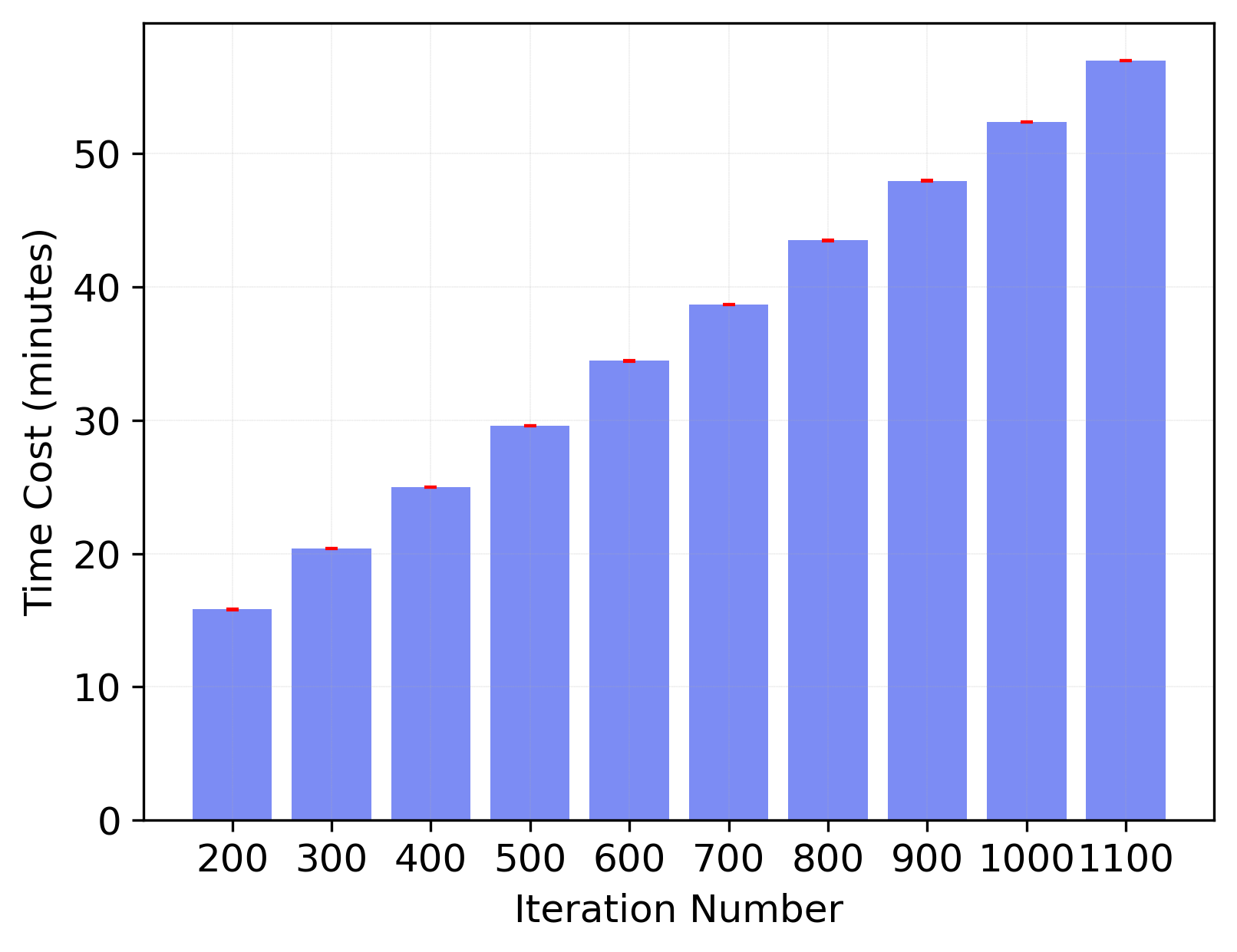}
\quad}
\caption{Quantization results of ViT-S in 3-bit, 4-bit, and 6-bit bit-width with increasing iteration number. (a) The Top-1 (\%) accuracy and (b) the time cost (minutes) is adopted as the evaluation metric for performance and efficiency respectively. We test three times to compute the mean and std.}
\label{iteration}
\end{figure}
\subsubsection{Effect of POS} Secondly, we study the influence of the progressive optimization strategy on the performance of low-bit quantization. To this end, in Table \ref{POS}, we report the accuracy of the model under different quantization bit-width using the conventional one-stage method and the proposed POS respectively. It is worth noting that the performance gain brought by the proposed POS is more significant in extremely low-precision quantization such as 4.41\% and 2.28\% in 3-bit and 4-bit quantized ViT-S respectively. For 6-bit quantization, although not obvious, our method still achieves 0.18\% higher Top-1 accuracy improvement compared with the conventional one-stage reconstruction method. Besides, as shown in Table \ref{ablation_study_summary}, the proposed POS is beneficial for both the block-wise reconstruction and PFCR, boosting their classification performance by 38.67\%. and 2.28\% severally. In particular, combining the POS and PFCR, the best Top-1 accuracy of 76.31\% is obtained, narrowing the performance gap to 5.08\% between the full-precision counterpart. These convincing results showcase that our proposed training strategy is suitable for the progressive reconstruction technique, providing a smoother and more stable optimization landscape for better results in low-bit quantized ViTs.

Furthermore, as demonstrated in Table \ref{timecost and top1}, the previous works like Outlier-Aware \cite{Outlier-awarema2024outlieraware} optimize the parameters of quantizers with 20000 iterations to achieve the satisfied performance, costing about 2 hours for reconstruction. In contrast, the proposed PFCR with POS takes only nearly 19 minutes with 300 base iterations to reconstruct the whole model, which saves the time requirement by 6.96 $\times$ and achieves 3.43\% performance improvement. It is worth noting that even the PFCR without POS obtains better classification performance with even lower time cost for reconstruction. As mentioned earlier in the proposed PFCR, the finer reconstructed units provide a better initialization for the coarser units, which contributes to the faster convergence speed, therefore striking a decent balance between quantization efficiency and performance.

However, as Table \ref{timecost and top1} demonstrates, the max memory requirement of our proposed method for the reconstruction process is 11.02 GB due to the coarser granularity during the progressive reconstruction process, which incurs the 6.37 GB additional hardware costs than the previous SOTA method Outlier-Aware. Regarding the significant performance improvement, we argue that this memory cost is acceptable and the quantization process is affordable for a single GPU to conduct the progressive reconstruction.


\begin{figure*}[t]
\centering
\includegraphics[width=0.82\textwidth]{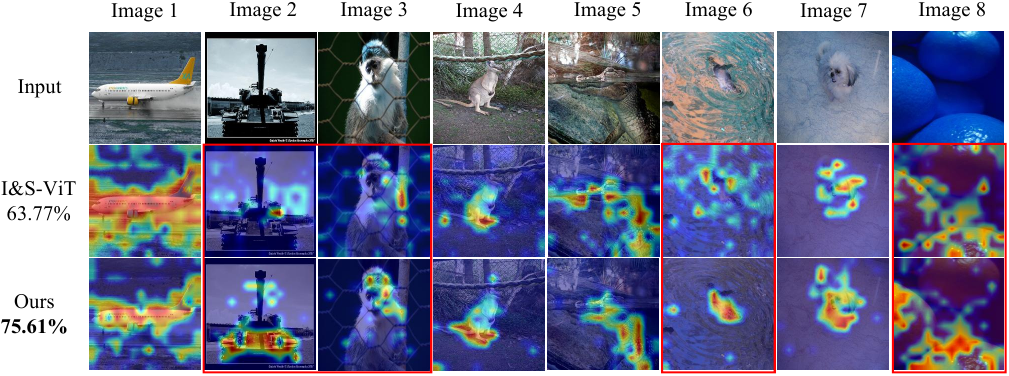} 
\caption{Qualitative visualization results of 3-bit quantized ViT-B using I\&S-ViT method and our proposed PFCR. We select 8 images from the validation set of ImageNet dataset and utilize the Grad-CAM \cite{GradCamselvaraju2017grad} technique to derive the attention maps.}
\label{visualization results}
\end{figure*}
\subsubsection{Effect of Iteration Number} Finally, we explore the effect of the iteration number on the quantization performance of the low-bit quantized ViT-S regardless of efficiency. Figure \ref{iteration}(a) illustrates the Top-1 accuracy of different low-bit quantized ViT-S with the increasing iteration number $iter_0$ from $200$ to $1100$. It is shown that the performance of 4-bit and 6-bit quantized models is not sensitive to the iteration number, where the longer reconstruction time does not bring obvious accuracy gains. Different from them, the 3-bit quantized model benefits significantly from the larger number of iterations. However, as shown in Figure \ref{iteration}(b), the quantization time grows linearly with the iteration number, reducing the efficiency of PTQ. Therefore, given these observations, we recommend that a more elaborate optimization process could be involved in extremely low-bit quantization like 3-bit when hardware resources are sufficient and high performance is required.

\subsection{Comparisons of Visualization Results}
To more intuitively demonstrate the effectiveness of our proposed method, we select 8 images from the ImageNet validation set and utilize the Grad-CAM \cite{GradCamselvaraju2017grad} to compare the visualization results with the block-wise I\&S-ViT \cite{IS-ViTzhong2023s} based on 3-bit quantized ViT-B in Figure \ref{visualization results}. As the figure shows, our proposed method pays better attention to the discriminative features such as Image 2 and Image 3. Besides, as Image 6 and Image 8 reveal, our quantized 3-bit ViT-B has a greater focus on informative objects, without distracting their attention from class-independent objects. Equipped with the progressive reconstruction technique, our method achieves a lower quantization error and better classification performance than the block-wise reconstruction works, contributing to the 11.84\% Top-1 accuracy improvement in 3-bit quantization.
\section{Conclusions}
In this paper, we find that the previous reconstruction methods in PTQ ignore the progressive relationship between different reconstruction granularity, which only predefines the granularity before optimization and derives the sub-optimal results in low-bit quantization settings. To this end, we propose progressive fine-to-coarse reconstruction (PFCR), a novel reconstruction method with mixed granularity in PTQ for accurate low-bit quantized vision transformers. PFCR applies the unified paradigm of progressive fine-to-coarse reconstruction, where the finer reconstructed units provide better initialization for coarser units, and the coarser granularity in turn finetunes the finer units. Specifically, PFCR fully utilizes the progressive relationships between different reconstruction granularity: the fine granularity is the building component for the coarser granularity. When two units are reconstructed, the coarser units are combined by them and begin to reconstruct. With this process iteratively, we complete the whole reconstruction in the end. Besides, we propose a two-stage optimization method POS, which alleviates the training difficulties in low-bit quantization and boosts the performance further. Experimental results show that our proposed method achieves state-of-the-art performance in low-bit quantized ViTs on different vision tasks including image classification, object detection, and instance segmentation, especially for extremely low bit-width like 3-bit and 4-bit.

The proposed progressive fine-to-coarse reconstruction method achieves the state-of-the-art performance on different datasets and various backbones for low-bit quantization in vision transformers. However, the coarser reconstruction units also incur additional memory requirements as shown in Table \ref{timecost and top1}, which may prevent efficient post-training quantization in some resource-limited environments. Besides, the effectiveness of our proposed method is under-explored in the large language models which are based on the transformer architecture. In the future, we will explore the more efficient implementation of our proposed method. For example, we can only optimize the quantization parameters in the coarser reconstruction granularity which saves the memory requirements significantly. Furthermore, for quantizing the large language models, we can explore the combination of our method with the parameter-efficient-fintuning technique for effective model quantization and specific task adaption.

\printcredits

\section*{Declaration of competing interest}
The authors declare that they have no known competing financial interests or personal relationships that could have appeared to influence the work reported in this paper.
\section*{Data availability}
Data will be made available on request.


\bibliographystyle{cas-model2-names}
\bibliography{own.bib}



\end{document}